\documentclass[10pt,twocolumn,letterpaper]{article}

\usepackage[pagenumbers]{cvpr} 

\usepackage{amsfonts}
\usepackage{amssymb}
\usepackage{pifont}
\usepackage{multirow} 
\usepackage{xcolor}
\usepackage{graphicx}
\usepackage{tikz}
\usepackage{float}

\definecolor{cvprblue}{rgb}{0.21,0.49,0.74}
\usepackage[pagebackref,breaklinks,colorlinks,allcolors=cvprblue]{hyperref}

\def\paperID{1001} 
\def\confName{CVPR}
\def\confYear{2026}

\title{FR-TTS: Test-Time Scaling for NTP-based Image Generation with Effective Filling-based Reward Signal}

\author{Hang Xu$^{1}$  \quad Linjiang Huang$^{2}$ \quad  \quad Feng Zhao$^{1}$\thanks{Corresponding author}\\
$^{1}$MoE Key Lab of BIPC, USTC $^{2}$Beihang University\\
}

\begin{document}

\maketitle
\begin{abstract}
Test-time scaling (TTS) has become a prevalent technique in image generation, significantly boosting output quality by expanding the number of parallel samples and filtering them using pre-trained reward models. However, applying this powerful methodology to the next-token prediction (NTP) paradigm remains challenging. The primary obstacle is the low correlation between the reward of an image decoded from an intermediate token sequence and the reward of the fully generated image. Consequently, these incomplete intermediate representations prove to be poor indicators for guiding the pruning direction, a limitation that stems from their inherent incompleteness in scale or semantic content.
To effectively address this critical issue, we introduce the Filling-Based Reward (FR). This novel design estimates the approximate future trajectory of an intermediate sample by finding and applying a reasonable filling scheme to complete the sequence. Both the correlation coefficient between rewards of intermediate samples and final samples, as well as multiple intrinsic signals like token confidence, indicate that  the FR provides an excellent and reliable metric for accurately evaluating the quality of intermediate samples.
Building upon this foundation, we propose FR-TTS, a sophisticated scaling strategy. FR-TTS efficiently searches for good filling schemes and incorporates a diversity reward with a dynamic weighting schedule to achieve a balanced and comprehensive evaluation of intermediate samples. We experimentally validate the superiority of FR-TTS over multiple established benchmarks and various reward models. Code is available at \href{https://github.com/xuhang07/FR-TTS}{https://github.com/xuhang07/FR-TTS}.
\end{abstract}    
\section{Introduction}
\label{sec:intro}
Test-time scaling (TTS) seeks to enhance output quality by utilizing a reward model for evaluation and subsequent filtering. Recent advancements~\cite{chen2025tts,singhcode,kim2025inference} in TTS have led to the development of methods that proactively integrate the reward model during the intermediate generation steps. This approach allows for a more efficient identification and selection of desirable outputs, a strategy that has been empirically validated and widely implemented in LLMs.

The adoption of TTS is anchored in generative paradigms like diffusion or flow matching~\cite{ma2025scaling}. While these models easily accommodate basic scaling (e.g., Best-of-N sampling), their iterative nature offers a superior advantage: the ability to evaluate and prune intermediate steps. Crucially, these intermediate samples preserve the same spatial scale and robustly approximate the final image's semantic content. Even without high-frequency details, reward signals from an external model are highly confident, and critically demonstrate a stable relative reward ranking. A sample with a high intermediate reward is thus highly likely to yield a final, superior-reward image~\cite{singhcode}. This scaling reliability is paramount, allowing the model to leverage intermediate evaluations for pre-emptive pruning and efficiently steer computation toward high-reward regions.

However, applying TTS in NTP autoregressive generation~\cite{vaswani2017attention} remains challenging. NTP-based models can only obtain a portion of the complete image during generation, and their reward cannot accurately represent the reward of the complete image. As we analyze in Sec. \ref{sec:failure}, existing evaluation methods are all inappropriate: cropping the generated portion results in incomplete scale (Fig.~\ref{fig:reward_calculate} (1)); filling ungenerated tokens with zero introduces irrelevant information (Fig.~\ref{fig:reward_calculate} (2)); and awaiting the final image is too resource-intensive (Fig.~\ref{fig:reward_calculate} (3)). In stark contrast, other autoregressive paradigms~\cite{tian2024visual,han2025infinity,li2024autoregressive} follow a coarse-to-fine order where scale remains constant and sufficient semantic information exists early for a high-confidence reward. Therefore, a new design for calculating intermediate rewards for NTP is critically needed.

To address this critical issue, we design a novel Filling-Based Reward (FR) mechanism (Fig.~\ref{fig:reward_calculate} (4)). This mechanism ensures both scale completeness and the presence of sufficient semantic information by randomly filling the ungenerated tokens using tokens already generated within the same sample. This strategy offers two key advantages: it preserves scale completeness by inherently avoiding image cropping, and it isolates relevant semantic information by duplicating the sample's own content, thereby eliminating external, irrelevant padding tokens. However, relying on a single random filling attempt cannot guarantee the generation of a plausible image for reliable reward assessment.

Thus, we generate multiple random filling schemes and assess the plausibility of each filling scheme with reward models. We then set the reward for the intermediate sample to be the maximum reward obtained from these filling schemes, which implicitly signifies the most plausible filling scheme. We demonstrate that the upper bound of the Filling-based Reward (FR) serves as a robust proxy for evaluating intermediate sample quality, as shown in Fig. \ref{fig:fr_illustration}. First, the correlation score between the FR of the intermediate sample and the complete sample’s reward demonstrates significant superiority over other reward calculation methods across all steps, as shown in Fig. \ref{fig:corr_score_2}. This implies that a higher FR for the intermediate sample correlates with a greater likelihood of achieving a higher reward upon full generation. Second,  FR’s effectiveness is further verified based on the discussion in ScalingAR~\cite{chen2025go}, where the FR upper bound exhibits strong concordance with the intrinsic statistics of AR models introduced by ScalingAR, including attention entropy in Fig. \ref{fig:attention_entropy}, Unified Score of endogenous signals in Fig. \ref{fig:inter_score} and token confidence in Fig. \ref{fig:confidence}, which indicates that FR helps the models generate with greater confidence and greater consistency. Consequently, we reframe the evaluation of intermediate samples as a search optimization problem targeting the upper bound of the FR.

Building upon the FR, we propose FR-TTS (Fig. \ref{fig:main_fig}), an efficient test-time scaling strategy for NTP-based autoregressive generative models defined by three core principles:
1. Efficient search for the upper bound of the FR: We redesign a coarse-to-fine search strategy, initiating a broad search followed by refinement via neighborhood search to efficiently locate high-quality filling schemes.
2. Diversity reward as a supplement: A diversity score is introduced to disperse generation trajectories, acting as a crucial supplement, especially in early steps where FR confidence is low.
3. Unified reward with dynamic weighting schedule: The FR weight is gradually increased with generation progress. Conversely, the diversity weight is increased when FR variance is low, actively enhancing sample differentiation.

To summarize, we propose FR-TTS, a TTS strategy to successfully utilize pre-trained reward models for NTP-based models. It leverages filling-based reward to obtain high-confidence signals during generation, thereby providing effective guidance for generation directions and significantly boosting the performance of TTS. We conduct experiments on multiple benchmarks, validating its superiority. 

\section{Related Work}
\label{sec:related}

\subsection{Test-Time Scaling in T2I Generation}
Test-Time Scaling (TTS) aims to improve the inherent performance of a model by increasing the computational cost during inference~\cite{muennighoff2025s1,snell2024scaling}. Unlike methods such as Chain of Thought~\cite{wei2022chain,feng2023towards}, which serialize multiple inference processes, TTS obtains the highest-reward output by parallelizing multiple samples and incorporating mechanisms for screening and pruning. TTS also has broad applications in T2I (Text-to-Image) generation tasks~\cite{li2024derivative,singhcode}, including flow matching~\cite{esser2024scaling} and autoregressive modeling~\cite{lee2022autoregressive,li2024autoregressive}.

To align with human preferences, an external, pre-trained reward model~\cite{xu2023imagereward,wu2023human,ma2025hpsv3} is often introduced in TTS. For instance, the best-of-N algorithm feeds multiple generated samples into the reward model and returns the sample with the highest reward. Some algorithms~\cite{ma2025hpsv3} go a step further, evaluating the intermediate samples and performing early pruning to ensure that the generation trajectory remains as close as possible to the high-reward region.

\subsection{Intermediate Evaluation in T2I Generation}
To better screen high-reward samples, many TTS strategies evaluate and prune the trajectory early in the generation process. In diffusion models, intermediate samples have already generated relatively rich information at the layout and semantic levels. Therefore, algorithms such as CoDe~\cite{singhcode} and SoP~\cite{ma2025scaling} evaluate intermediate samples with reward models, and then obtain samples for the next step through screening methods such as importance sampling.

Similar approaches have also been introduced in autoregressive generation, such as TTS-VAR~\cite{chen2025tts} for next-scale prediction. However, for many works involving next-token prediction (NTP), it is challenging to incorporate a pre-trained reward model, given the incompleteness of the intermediate samples at both the scale and semantic levels. Therefore, ScalingAR~\cite{chen2025go} chooses to circumvent this issue by considering the inherent properties of the NTP model itself to measure the quality of the generated trajectory, but this does not explicitly align with human preferences.

\section{Analysis for TTS in NTP-based Generation}
\label{sec:analysis}

\subsection{Preliminaries}
\label{sec:preliminary}
\paragraph{NTP-based autoregressive modeling.} In NTP-based models, images are modeled as sequences of tokens, where the already generated tokens act as the condition to predict the next token~\cite{vaswani2017attention}:
\begin{equation}
    p(x_1,x_2,\ldots,x_T)=\prod_{t=1}^Tp(x_t|x_1,x_2,\ldots,x_{t-1}).
\end{equation}
Unlike other generative paradigms (e.g., mask modeling~\cite{li2024autoregressive}, flow matching~\cite{esser2024scaling}), where the image retains its full scale and semantics during the generation process and details are progressively refined, NTP-based models fail to adequately preserve either the full scale or semantics before the complete image sequence is obtained.

\begin{figure}[t]
   \centering
    \subfloat[Reward calculate methods.\label{fig:reward_calculate}]{
        \includegraphics[width=0.95\linewidth]{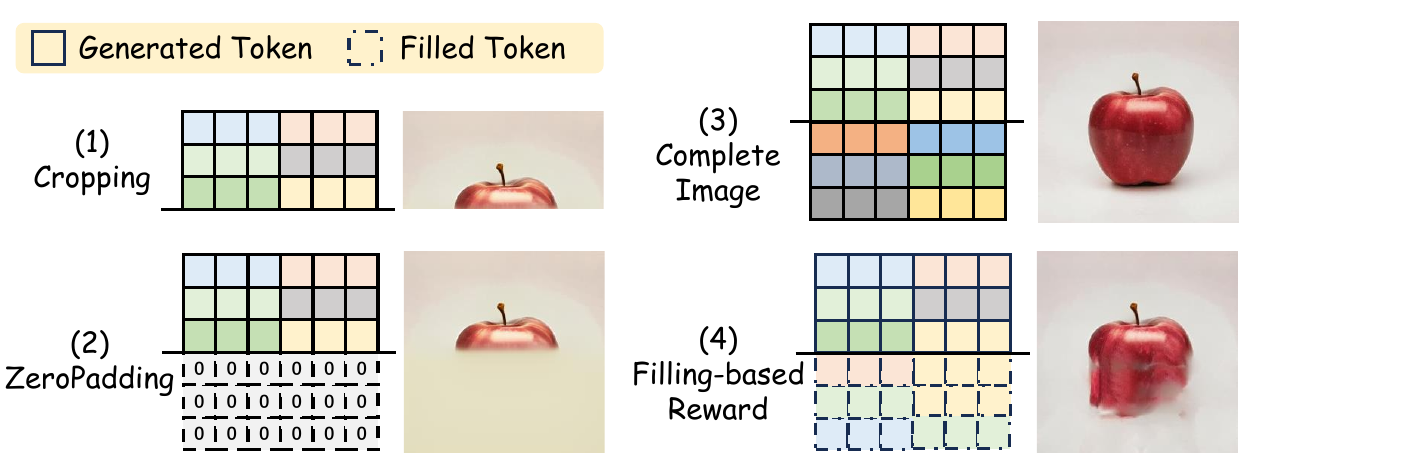}
    }
    \\
    \subfloat[Correlation score of different models.\label{fig:corr_score}]{
        \includegraphics[width=0.85\linewidth]{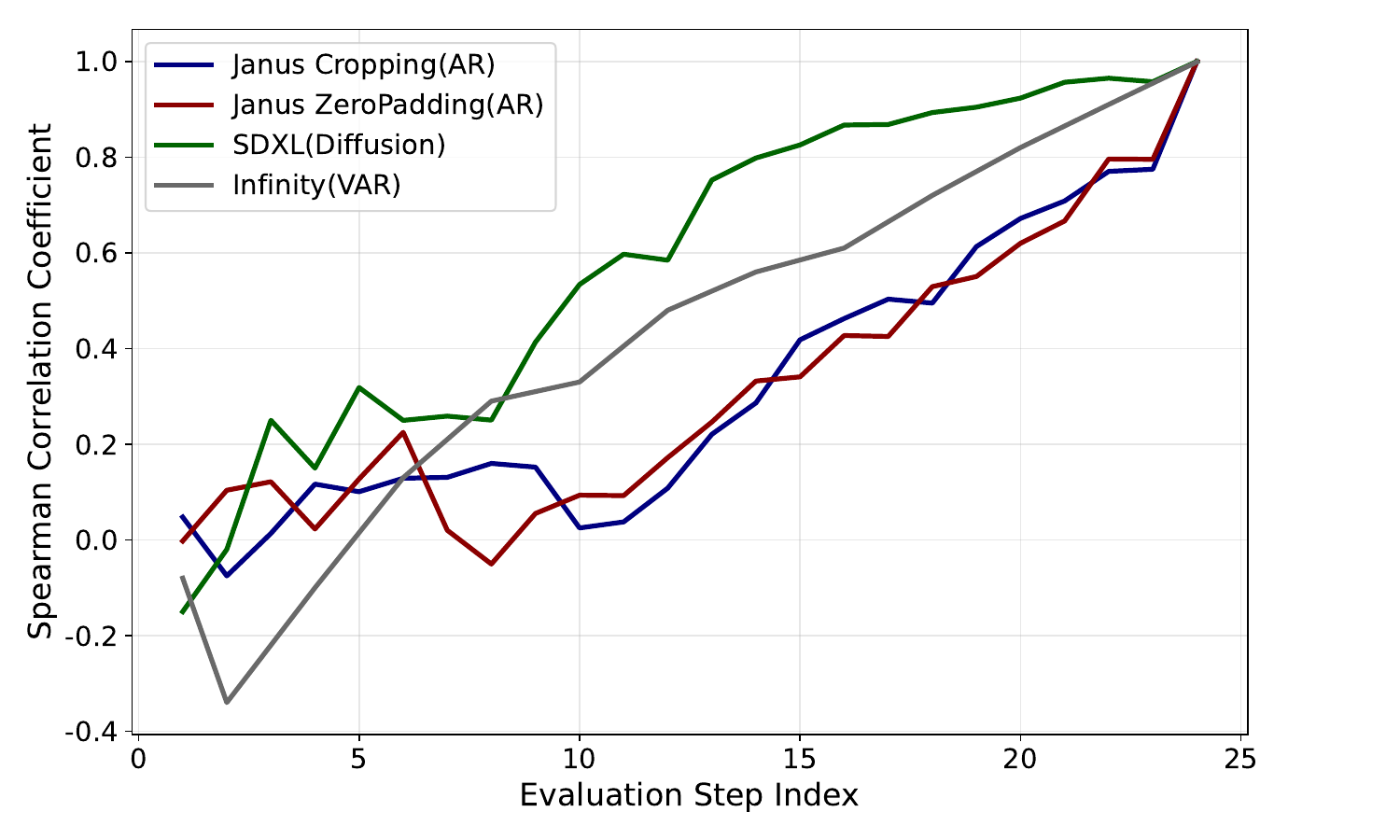}
    }
    \caption{\textbf{Comparison of the Spearman \cite{wissler1905spearman} correlation scores between the reward of intermediate samples and the reward of final complete ones for different paradigms.} For the NTP paradigm, we used two different reward calculation methods: Cropping the already generated part, ZeroPadding the ungenerated part, as shown in (a). The results in (b) show that the NTP paradigm's correlation score to the last step's is lower at all steps compared to other paradigms (e.g., Diffusion and VAR~\cite{tian2024visual}), where we evaluate 24 steps at equal intervals (13 steps evaluated at Infinity, which are equivalently mapped to the 24 steps).}
\end{figure}

\begin{figure*}[t]
    \centering
    \subfloat[Correlation score of different methods.\label{fig:corr_score_2}]{
        \includegraphics[width=0.23\linewidth]{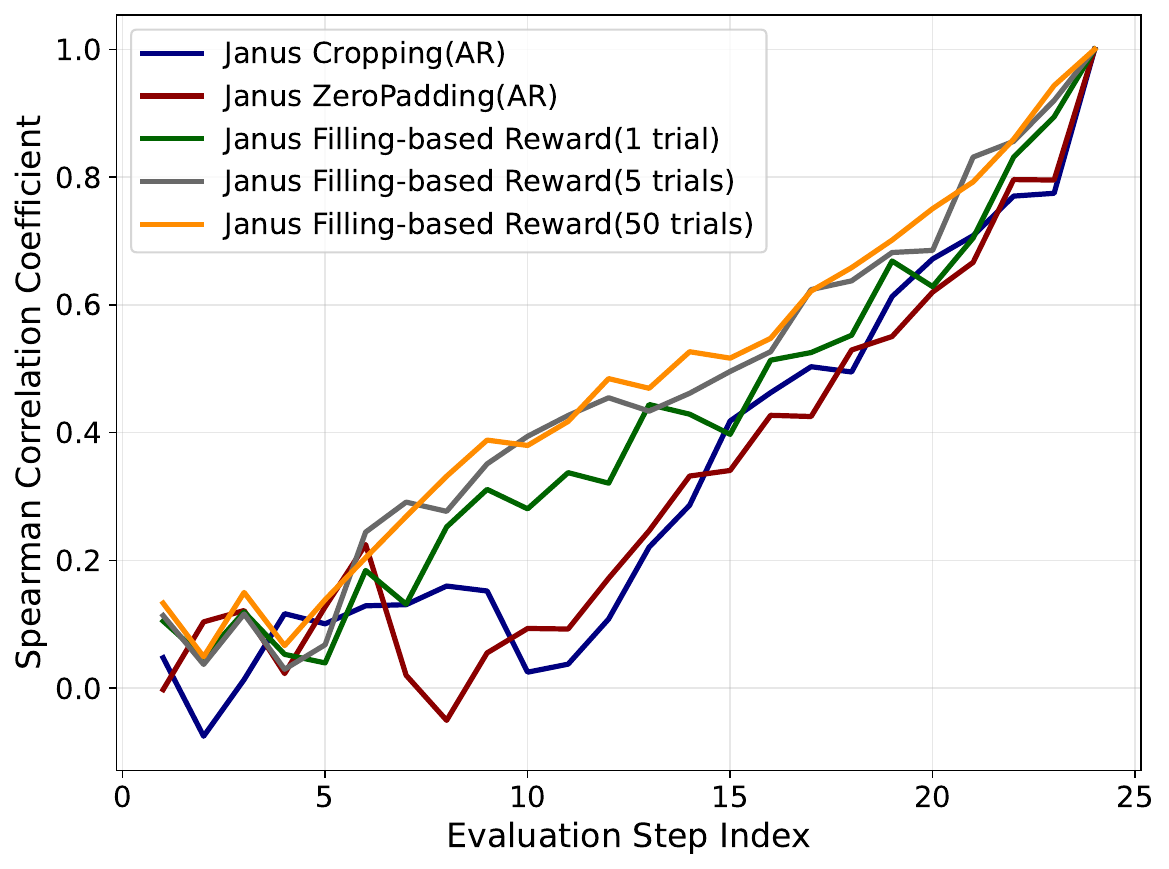}
    }
    \hfill
    \subfloat[Attention entropy of filling trials with high and low reward. \label{fig:attention_entropy}]{
        \includegraphics[width=0.24\linewidth]{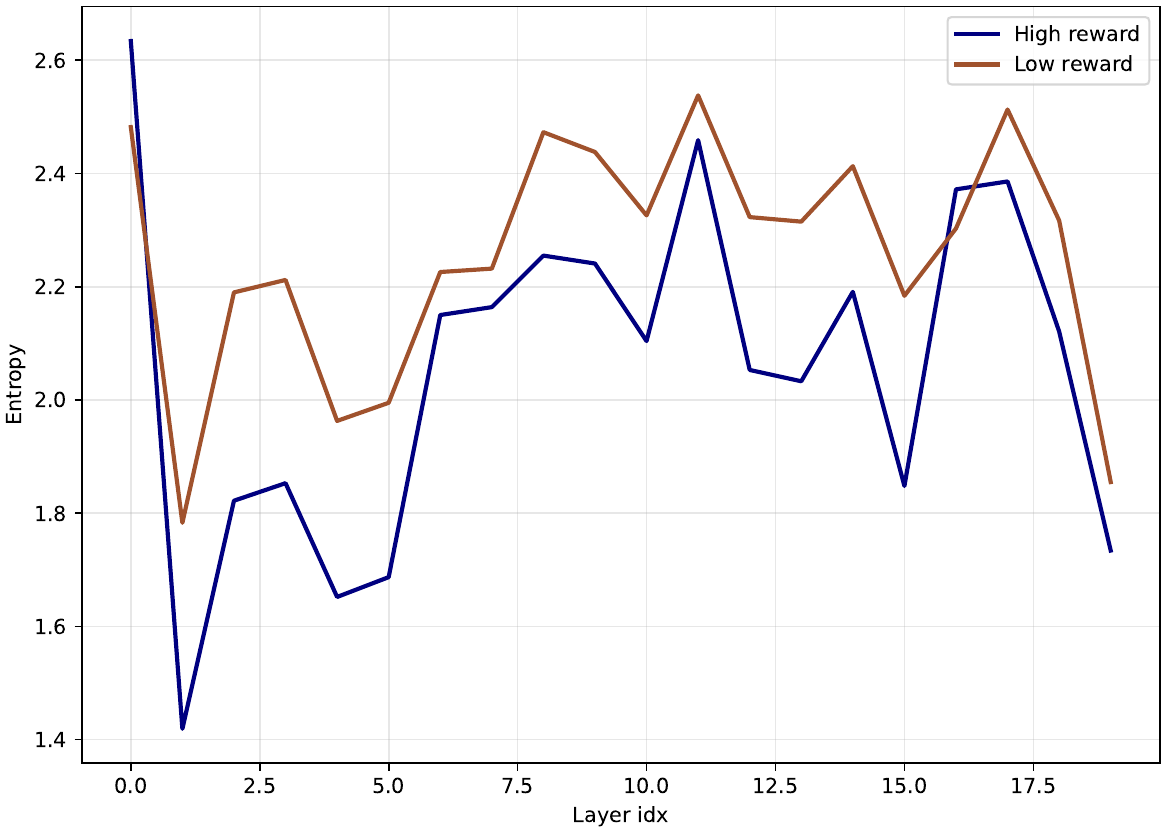}
    }
    \hfill
    \subfloat[Endogenous signals of generated tokens. \label{fig:inter_score}]{
        \includegraphics[width=0.23\linewidth]{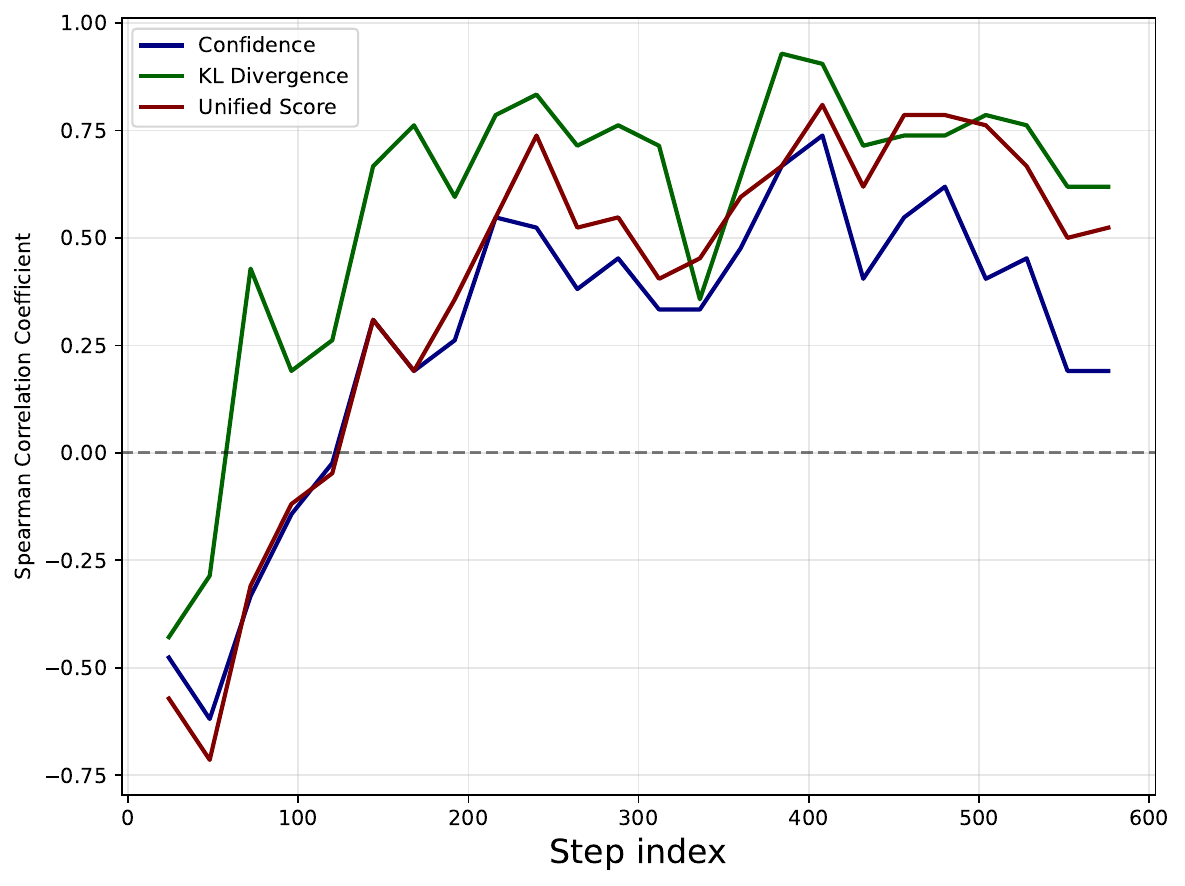}
    }
    \hfill
    \subfloat[Confidence of ungenerated tokens. \label{fig:confidence}]{
        \includegraphics[width=0.23\linewidth]{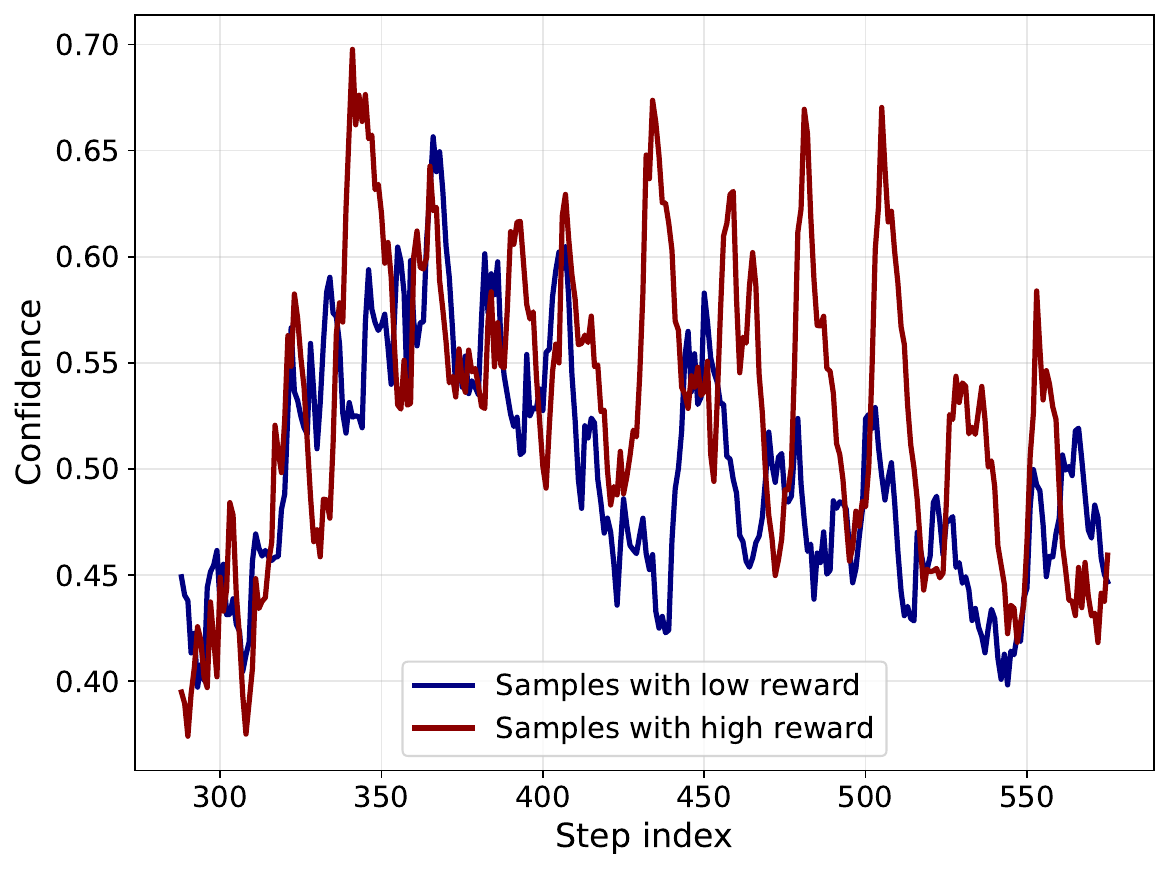}
    }
    
    \caption{\textbf{We illustrate four aspects to show that the Filling-based reward (FR) is a highly effective metric for evaluating intermediate samples. We conduct experiments on Janus pro 7B~\cite{chen2025janus} with 7K+ prompts of Open-Image-Preferences-v1~\cite{OpenImagePreferencesV1Blog}.}(a) FR exhibits a higher correlation score, and the score continues to increase as the number of random filling times increases. (b) The filling scheme with a higher reward consistently shows lower entropy in the attention maps across different layers, which indicates that its semantics are more concentrated and less diffused \cite{ma2025towards,chen2025go}. (c) After several steps, there is a high correlation between the endogenous signal scores introduced by ScalingAR \cite{chen2025go} of the generated tokens and FR. (d) The sample with a higher FR has a higher confidence on the ungenerated tokens.}
    \label{fig:fr_illustration}
\end{figure*}
\subsection{Failure of direct evaluation with reward models}
\label{sec:failure}
The utilization of external reward models to evaluate the trajectory during generation can effectively guide the process and enhance generation quality, a finding verified in many paradigms. Nonetheless, in NTP-based models, this method seldom results in metric gains, as shown by ScalingAR~\cite{chen2025go}. The superficial reason is that the intermediate reward cannot effectively reflect the relative quality of the samples.
\paragraph{Correlation score for measuring the preservation of the relative relationship of sample rewards.} We utilize the correlation coefficient~\cite{wissler1905spearman} between the intermediate rewards and the true reward to measure the extent to which the relative relationship of the rewards is preserved during the generation process. For NTP-based models, we perform the evaluation after generating an entire row to reduce the number of evaluations. For other paradigms, we control the number of steps to be in a similar range and perform the evaluation after every generation step. Furthermore, considering the incompleteness of the generation, we consider two evaluation methods: cropping the already generated portion (Cropping) and filling the ungenerated token IDs with 0 (ZeroPadding), as shown in Fig. \ref{fig:reward_calculate}. 

\paragraph{Results and analysis.} As shown in Fig. \ref{fig:corr_score}, the correlation scores for Flux(flow matching)~\cite{flux2024} and Infinity(next-scale prediction)~\cite{han2025infinity} are consistently and significantly higher than that of Janus-pro(NTP) across all steps. Additionally, Cropping leads to a worse correlation. Assuming the already generated tokens are $[x_1, x_2, \ldots, x_t]$, the intermediate reward at time $t$ is defined as:
\begin{equation}
R_t^{\text{cropping}} = P\{R \mid x_1, x_2, \ldots, x_t\}.
\end{equation}
It is worth noting that reward models often accept images with a fixed aspect ratio. This effectively requires first resizing the cropped image to a consistent scale, resulting in a new token sequence:
\begin{equation}
    R_t^{\text{cropping}} = P\{R \mid x'_1, x'_2, \ldots, x'_T\}.
\end{equation}
The true reward is obtained based on the complete sequence of tokens: 
\begin{equation}
R^{\text{true}} = P\{R \mid x_1, x_2, \ldots, x_T\}.
\end{equation}
Since VQ encoding is extremely sensitive to augmentation~\cite{yue2025understand}, the tokens $x'_i$ and $x_i$ become almost entirely different, which consequently leads to a very low correlation between $R_t^{\text{cropping}}$ and $R^{\text{true}}$. In contrast, $R_t^{\text{ZeroPadding}}$ fills all ungenerated token IDs with a default value of $0$:
\begin{equation}
R_t^{\text{ZeroPadding}} = P\{R \mid x_1, x_2, \ldots,x_t, 0, \ldots, 0\}.
\end{equation}
This ensures consistency in scale and simultaneously guarantees that the prefix of the condition remains consistent, thus improving the correlation score. However, because this meaningless information interferes with the reward model's judgment, the resulting correlation score is still significantly lower than that of other paradigms.

\subsection{Filling-based Reward for Test-time Scaling}
\label{sec:fr}
Since preserving the scale is necessary to keep the prefix tokens unchanged, and we also want to introduce relevant semantic information into the ungenerated part, the most suitable approach is to fill the ungenerated part with the already generated portion. We call this Filling-based Reward (\textbf{FR}). However, random filling can easily lead to chaotic layouts and details. To measure the efficacy of different filling schemes, we directly decode the scheme and then calculate the reward using the reward model. Theoretically, a filling scheme with a high reward inherently possesses content that is relatively harmonious and consistent. Therefore, we repeatedly generate random filling schemes and select the highest reward among them to serve as the intermediate reward. We observed an interesting fact: as the number of random attempts increases, the reward of the final generated sample also increases. This indicates that the upper bound of the filling-based reward is an excellent metric for evaluating the generation trajectory, and what we are doing is essentially searching for this upper bound. To further validate this perspective, To further validate this perspective, we leverage the correlation score and several conclusions from ScalingAR~\cite{chen2025go}, as shown in Fig. \ref{fig:fr_illustration}.

\paragraph{Higher correlation score.} As shown in Fig. \ref{fig:corr_score_2}, the FR clearly has a higher correlation score compared to Cropping and ZeroPadding. Specifically, the result from a single random attempt is already close to that of ZeroPadding, and as the number of random attempts increases, the correlation score of the filling-based reward continues to improve.

\paragraph{More focused attention of the complete token sequence.} We feed the complete random scheme into Janus-pro and obtain its attention map. We calculate the average attention entropy for each layer across the token dimension:
\begin{equation}
    \bar{E}^l_{attn} = - \frac{1}{T} \sum_{i=1}^{T} \left( \sum_{j=1}^{T} A^l_{i, j} \log_2 (A^l_{i, j}) \right)
\end{equation}
Where $l$ represents the attention map of the $l$-th layer, and $A_{i,j}$ represents the attention weight between the $i$-th and $j$-th tokens. As shown in Fig. \ref{fig:attention_entropy}, better filling schemes tend to have lower attention entropy, which reflects that filling schemes with higher rewards possess more centralized semantics and more harmonious content~\cite{ma2025towards}.
\begin{figure*}[t]
	\centering
	\includegraphics[width=0.95\linewidth]{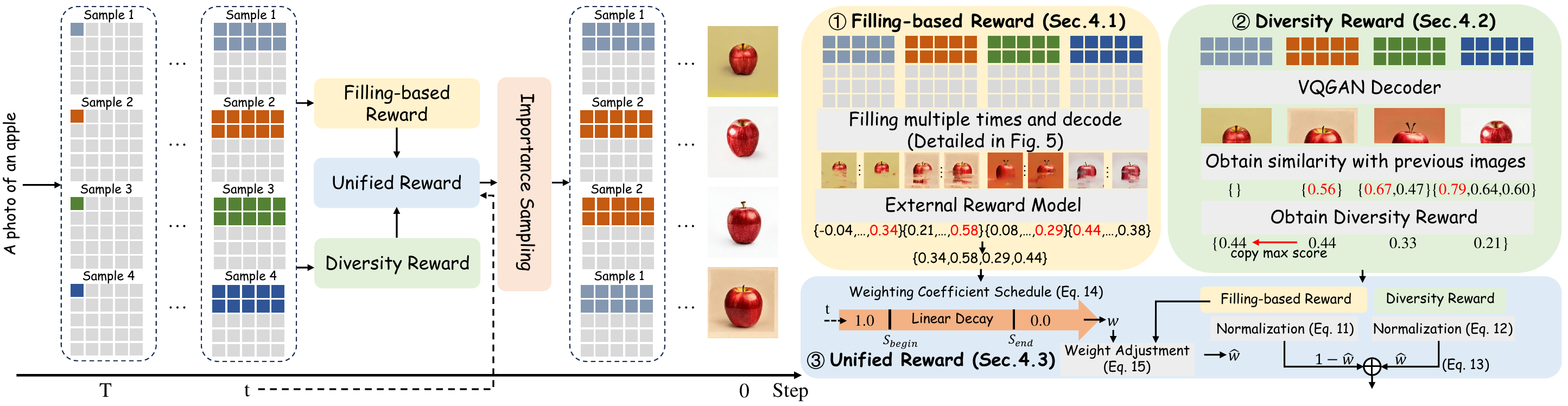}
	\caption{\textbf{Our proposed FR-TTS.} Our scaling strategy is built upon three pivotal design principles that ensure efficiency and robust search: \textcircled{1} Efficient Search for the Upper Bound of Filling-based Reward (FR): We generate filling schemes multiple times to search for the upper bound of FR (we takes two best searches in middle steps as an example here), where details are shown in Fig. \ref{fig:fr}. \textcircled{2} Diversity Reward: To encourage broader exploration, we assign a diversity reward to each sample, calculated as one minus its maximum similarity score relative to previously generated samples. \textcircled{3} Unified Reward: Based on the increasing correlation score of the FR over time in Fig. \ref{fig:corr_score_2}, we employ a dynamic weighting coefficient schedule. We further incorporate variance-based adjustments to the FR weights, allowing for enhanced differentiation in subsequent steps. Finally, we utilize Importance Sampling based on the unified reward to obtain new parallel samples.} 
	\label{fig:main_fig}
\end{figure*}

\begin{figure}[h]
	\centering
	\includegraphics[width=0.95\linewidth]{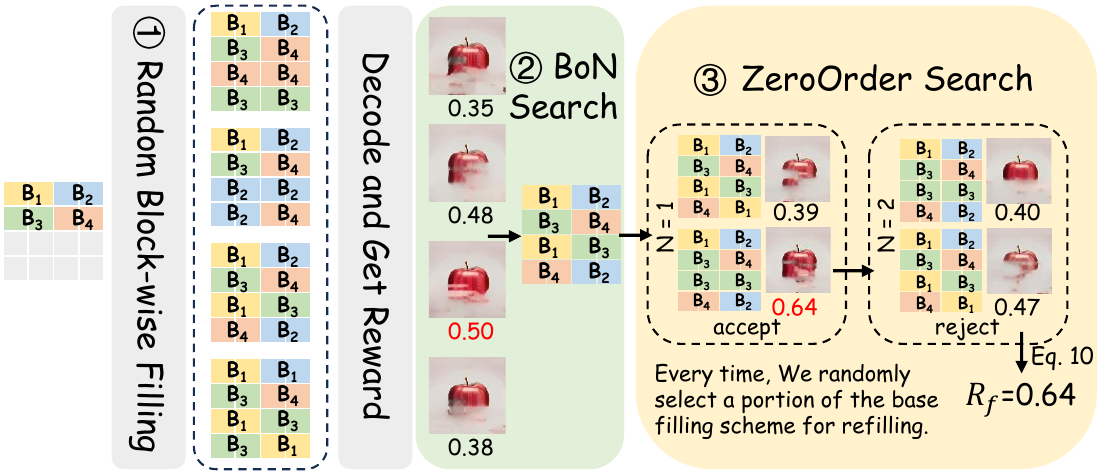}
	\caption{\textbf{Coarse-to-Fine Search Strategy for Efficient FR Scaling.} Our approach for efficiently finding the upper bound of the Filling-based Reward (FR) utilizes a coarse-to-fine search strategy. Initially, we perform multiple block-wise random fillings to establish a base filling scheme via BoN selection. Building upon this base, we transition to Zero-Order~\cite{ma2025scaling} optimization, where we iteratively refill a small number of its blocks. If this local refinement yields a higher reward score, the new block configuration replaces the current base filling scheme; otherwise, the replacement is rejected. The reward of the final optimized base filling scheme then serves as the final FR for the intermediate sample.} 
	\label{fig:fr}
\end{figure}
\paragraph{Better endogenous signals of generated tokens.} For the FR across different samples, a higher reward often corresponds to a higher endogenous signal score for the already generated tokens, as shown in Fig. \ref{fig:inter_score}. The endogenous signal score is calculated following ScalingAR~\cite{chen2025go}, which reflects the confidence of the image tokens and their utilization of the text condition. Specifically, the correlation coefficient between our reward and the text utilization is higher, which indicates that our reward better reflects the semantic properties of the generated portion.

\paragraph{Higher confidence of ungenerated tokens.} Samples with a higher FR tend to have higher token confidence during the subsequent generation process, as shown in Fig. \ref{fig:confidence}, which implicitly aligns with the motivation of ScalingAR. This demonstrates that our reward helps guide the model toward a more stable direction.

From these four aspects, we have demonstrated that the upper bound of the FR is an excellent evaluation metric, thus successfully transforming the problem of evaluating the generation trajectory into a problem of searching for the reward upper bound. In the next section, we further discuss how to search efficiently.
\section{FR-TTS}
\label{sec:method}
This section is devoted to detailing FR-TTS, a test-time scaling strategy specifically designed for NTP-based models, as shown in Fig. \ref{fig:main_fig}. The strategy is centered around three key components: the filling-based reward, the diversity reward, and a unified reward that integrates the former two. We will introduce these three components sequentially across three separate subsections.

\subsection{Efficiently Scaling for the Filling-based Reward}
Although the upper bound of the filling-based reward can effectively evaluate the intermediate samples, randomly attempting it many times is inefficient, and arguably less efficient than directly generating complete images. Therefore, to approach this upper bound efficiently, we introduce the following two designs, as shown in Fig. \ref{fig:main_fig}\textcircled{1} and Fig. \ref{fig:fr}.
\begin{table*}[t]
\centering

\caption{\textbf{Quantitative evaluations of general text-to-image generation on Open-Image-Pref-V1 prompts.} This study compares the strengths and weaknesses of different scaling strategies based on commonly used reward models: HPSv3, ImageReward, ClipScore, AestheticScore, and HPSv2. Furthermore, we analyze the time consumption associated with each strategy.}

\begin{tabular}{ccccccc}
\toprule
Methods  & HPSv3$\uparrow$ & ImageReward$\uparrow$ & CLIPScore$\uparrow$ &AestheticScore$\uparrow$ &HPSv2$\uparrow$ &Consumption$\downarrow$\\ \hline

Janus-pro 1B  &10.35 &1.23 &0.83 &5.49 &0.254 &\textbf{10s/img} \\ 
+Best-of-N  &12.22 &1.34 &0.85 &5.66 &0.279 &12s/img \\
+\textbf{Ours}  &\textbf{13.24} &\textbf{1.45} &\textbf{0.93} &\textbf{5.82} &\textbf{0.288} &18s/img \\ \hline
Janus-pro 7B  &12.55 &1.33 &0.95 &5.52 &0.280 &\textbf{17s/img}\\
+Best-of-N  &13.08 &1.47 &1.04 &5.71 &0.293 &25s/img\\
+\textbf{Ours}  &\textbf{13.69} &\textbf{1.68} &\textbf{1.18} &\textbf{5.94} &\textbf{0.302} &31s/img\\

\bottomrule
\end{tabular}

\label{tab:main_tab}
\end{table*}
\paragraph{Block-wise filling.} Some previous work has shown that filling or matching needs to be performed using larger units~\cite{efros2023image,barnes2009patchmatch}. This ensures continuous and consistent semantics within the block while significantly reducing the likelihood of chaotic filling schemes. Assuming the complete sequence is $X = (x_1, x_2, \ldots, x_t)$, where $t$ is the sequence length. We segment the already generated sequence $X$ into $m$ non-overlapping blocks $B_m$:
\begin{equation}
    X = B_1 \cup B_2 \cup \ldots \cup B_m
\end{equation}
Each block $B_i$ consists of $K$ consecutive tokens, where $K$ is the Block Size. We assume that $m$ blocks ($B_1$ through $B_m$) have already been generated, and we need to perform filling for the subsequent blocks, from $B_{m+1}$ up to $B_M$.The filling scheme for these subsequent blocks is defined by randomly selecting a block from the set of already generated blocks to serve as the content for the unfilled block:
\begin{equation}
    S = (B_{m+1},\dots,B_M)
\end{equation}
Where, for $m<i<=M$,we have:
\begin{equation}
    B_i = \text{Sample}(B_1, B_2, \ldots, B_m)
\end{equation}
Our objective is to find the highest reward of all filling schemes, as shown in Fig. \ref{fig:fr}\textcircled{1}, which is an optimization task aimed at maximizing the Filling-based Reward ($R_f$):
\begin{equation}
    R^*_f = \max_{S \in \mathcal{S}} R_f(X, S)
\end{equation}

\paragraph{Coarse-to-fine searching.} The search for the upper bound of the filling-based reward is a reward-based TTS task, and therefore, the introduction of an appropriate TTS strategy can also improve the search efficiency. The strategy of filling multiple times and taking the highest reward in Sec. \ref{sec:fr} can be seen as BoN. Considering that this infilling lacks sequential order, we introduce the ZeroOrder algorithm~\cite{ma2025scaling} and redesign a coarse-to-fine search strategy, as shown in Fig. \ref{fig:fr}\textcircled{2} and \textcircled{3}. Specifically, we first establish a base scheme with BoN and then use a small number of neighborhood searches to refine the base filling scheme.

\subsection{Diversity Reward for Trajectory Dispersion}
In the early steps, since the semantic information is too sparse (often just background), we introduce a diversity reward to ensure the generated trajectories are sufficiently dispersed in Fig. \ref{fig:main_fig}\textcircled{2}. Specifically, we decode and crop the already generated portion. Then, for the $i$-th sample, we calculate its distance from samples $0$ to $i-1$ with VGG~\cite{simonyan2014very}, and the minimum of these distances is taken as the $i$-th sample's similarity score, $R_s$. The diversity score is then expressed as $R_d = 1 - R_s$. For the first sample (sample $0$), as there are no preceding samples, we assign it the highest diversity score among all samples. In this way, we ensure the diversity of trajectories in the early stages of generation, providing more possibilities for subsequent filtering.

\subsection{Unified Reward with Weighting Schedule}
As shown in Fig. \ref{fig:main_fig}\textcircled{3}, we first perform normalization within each of the two rewards to standardize their scale:
\begin{equation}
    \hat{R_f} = \frac{R_f - \min(R_f)}{\max(R_f) - \min(R_f)}
\end{equation}
\begin{equation}
    \hat{R_d} = \frac{R_d - \min(R_d)}{\max(R_d) - \min(R_d)}
\end{equation}
Then we combine the two scores to obtain a unified reward:
\begin{equation}
    R_u = w\hat{R_d}+(1-w)\hat{R_f}
\end{equation}
To better combine the advantages of the two rewards, we set the weighting coefficient to gradually decrease throughout the generation process, where we define two cutoffs, $S_{\text{begin}}$ and $S_{\text{end}}$:
\begin{equation}
    w(t) = 
\begin{cases} 
1 & \text{if } t > S_{\text{begin}} \\
\frac{t - S_{\text{end}}}{S_{\text{begin}} - S_{\text{end}}} & \text{if } S_{\text{end}} \le t \le S_{\text{begin}} \\
0 & \text{if } t < S_{\text{end}} 
\end{cases}
\end{equation}
At the same time, we consider that if the variance of $R_f$ is too small, it likewise cannot effectively distinguish the quality of the samples. Therefore, we appropriately adjust $w(t)$ based on the variance of $R_f$:
\begin{equation}
    \hat{w} = w-\frac{1}{1+\exp[-(\mathbf{var}(R_f)-v_c)*v_s]}+0.5
\end{equation}
$v_c$ and $v_s$ are used to control the center and strength of the adjustment, respectively. Then we utilize $\hat{w}$ as the weighting coefficient to combine the two rewards.

\begin{table*}[t]
\centering
\caption{\textbf{Quantitative evaluations of compositional text-to-image generation on GenEval.}}

\scalebox{0.91}{
\begin{tabular}{cccccccccc}
\toprule
Methods & Arc. & One Obj. & Two Obj. & Count &Colors &Position &Color Attr. &Overall &ImageReward\\ \hline
SDXL &Diff. &0.98 &0.74 &0.39 &0.85 &0.15 &0.23 &0.55 &- \\
Flux &Diff. &0.98 &0.81 &0.74 &0.79 &0.22 &0.45 &0.66 &-\\
Infinity &VAR &1.00 &0.78 &0.60 &0.85 &0.25 & 0.55 &0.67 &- \\
LightGen &MAR &0.99 &0.65 &0.87 &0.60 &0.22 &0.43 &0.62 &-\\
Emu3 &AR(NTP) &0.98 &0.71 &0.34 &0.81 &0.17 &0.21 &0.54 &- \\
Show-o &Diff.+AR &0.98 &0.85 &0.67 &0.81 &0.28 &0.55 &0.69 &-\\ \hline
Janus-pro 1B &AR(NTP) &0.95 &0.67 &0.51 &0.81 &0.43 &0.45 &0.64 &-\\ 
+Best-of-N &AR(NTP) &0.98 &0.85 &0.56 &0.81 &0.58 &0.63 &0.73 &-\\
+\textbf{Ours} &AR(NTP) &\textbf{1.00} &\textbf{0.90} &\textbf{0.57} &\textbf{0.87} &\textbf{0.60} &\textbf{0.64} &\textbf{0.77} &- \\ \hline
Janus-pro 7B &AR(NTP) &0.96 &0.74 &0.58 &0.87 &0.56 &0.59 &0.72 &-\\
+Best-of-N &AR(NTP) &1.00 &0.88 &0.64 &0.88 &0.65 &0.71 &0.79 &-\\
+\textbf{Ours} &AR(NTP) &\textbf{1.00} &\textbf{0.94} &\textbf{0.69} &\textbf{0.90} &\textbf{0.76} &\textbf{0.75} &\textbf{0.84} &-\\ \hline
LlamaGen &AR(NTP) &0.72 &0.28 &0.21 &0.59 &0.04 &0.05 &0.30 &-0.38\\
+Best-of-N &AR(NTP) &0.85 &0.56 &0.32 &0.67 &0.09 &0.11 &0.42 &0.55\\
+ScalingAR &AR(NTP) &0.90 &0.56 &0.33 &0.67 &\textbf{0.15} &0.14 &0.47 &0.21\\
+\textbf{Ours} &AR(NTP) &\textbf{0.93} &\textbf{0.60} &\textbf{0.38} &\textbf{0.69} &0.14 &\textbf{0.14} &\textbf{0.50} &\textbf{0.73}\\

\bottomrule
\end{tabular}}

\label{tab:geneval}
\end{table*}
\begin{table}[t]
\centering

\caption{\textbf{Quantitative evaluations of compositional text-to-image generation on TIIF-Bench.}}

\scalebox{0.75}{
\begin{tabular}{cccccccc}
\toprule
Methods & \multicolumn{2}{c}{Basic} & \multicolumn{2}{c}{Advanced} & \multicolumn{2}{c}{Designer}  &Overall\\ 
&Long &Short &Long &Short &Long &Short & \\ \hline

Janus-pro 1B  &0.60 &0.47 &0.49 &0.38 &0.67 &0.66 &0.47\\ 
+Best-of-N  &0.64 &0.56 &0.56 &0.44 &0.65 &0.65 &0.53\\
+\textbf{Ours}  &\textbf{0.71} &\textbf{0.61} &\textbf{0.56} &\textbf{0.49} &\textbf{0.67} &\textbf{0.66} &\textbf{0.57}\\ \hline
Janus-pro 7B  &0.80 &0.74 &0.69 &0.59 &0.66 &0.67 &0.68\\
+Best-of-N  &0.81 &0.80 &0.68 &0.64 &\textbf{0.71} &0.60 &0.71\\
+\textbf{Ours}  &\textbf{0.84} &\textbf{0.83} &\textbf{0.69} &\textbf{0.67} &0.67 &\textbf{0.69} &\textbf{0.74}\\ \hline
LlamaGen &0.51 &0.58 &0.41 &0.44 &0.35 &0.34 &0.46\\
+Best-of-N &0.59 &0.66 &0.43 &0.46 &0.41 &0.37 &0.49\\
+ScalingAR &0.63 &0.68 &0.46 &0.49 &0.44 &\textbf{0.41} &0.52\\
+\textbf{Ours} &\textbf{0.64} &\textbf{0.72} &\textbf{0.49} &\textbf{0.54} &\textbf{0.46} &0.38 &\textbf{0.56} \\

\bottomrule
\end{tabular}}

\label{tab:tiif}

\end{table}
\section{Experiment}
\label{sec:exp}

\subsection{Experimental Settings}
\paragraph{Baselines.}
We conduct our experiments across three baselines: Janus pro 1B~\cite{chen2025janus}, Janus pro 7B~\cite{chen2025janus}, and LLamaGen~\cite{sun2024autoregressive}. We compare FR-TTS with Best-of-N (BoN) and ScalingAR~\cite{chen2025go}. For all strategies, we set the number of samples per step to 8. Specifically, for Janus-pro, the number of steps is 576 ($24 \times 24$); we use rows as the basic unit and perform a filtering step every time 4 rows are generated. For LlamaGen, the total number of steps is 1024 ($32 \times 32$); we similarly execute a filtering step after every 4 rows are generated. For other paradigms, we select representative models, including SDXL~\cite{podell2023sdxl}, Flux~\cite{flux2024}, Infinity~\cite{han2025infinity}, LightGen~\cite{wu2025lightgen}, Emu3~\cite{wang2024emu3} and show-o~\cite{xie2024show}.
\vspace{-5pt}
\begin{figure}[h]
	\centering
	\includegraphics[width=1\linewidth]{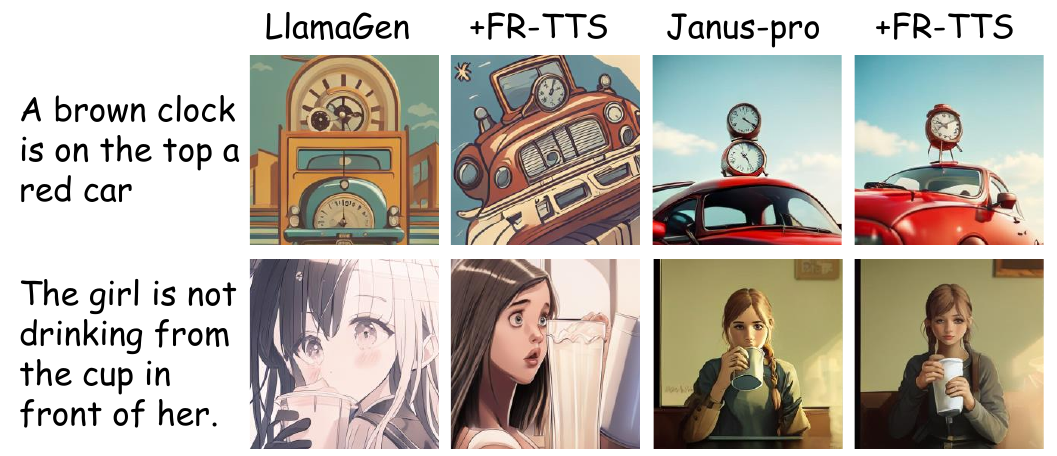}
	\caption{\textbf{Visual results of prompts from TIIF-Bench.}} 
    \vspace{-8pt}
	\label{fig:visual_results}
\end{figure}
\paragraph{Reward models and benchmarks.}
To measure image quality, we employ five pre-trained reward models: ImageReward~\cite{xu2023imagereward}, HPSv3~\cite{ma2025hpsv3}, ClipScore~\cite{hessel2021clipscore}, AestheticScore~\cite{Schuhmann:aesthetics}, and HPSv2~\cite{wu2023human}. We primarily aim to measure general generation capability; thus, we conduct our main experiments using the open-image-pref-v1 prompts~\cite{OpenImagePreferencesV1Blog}. Additionally, we conduct experiments on GenEval~\cite{ghosh2023geneval} and TIIF-Bench~\cite{wei2025tiif} for a more comprehensive comparison.

\subsection{Main Results}
\paragraph{Quantitative comparisons on reward models.} As shown in Tab. \ref{tab:main_tab}, our method consistently outperforms the conventional BoN strategy on both Janus-pro 1B and Janus-pro 7B. Simultaneously, while our method achieves a substantial performance gain, the resulting additional overhead is acceptable, where FR-TTS introduces a 50\% and 24\% increase in inference time on Janus Pro 1B and 7B compared with BoN, respectively. Here, a direct comparison with ScalingAR is inappropriate, considering that its objective is not to achieve better alignment with a reward model.

\paragraph{Quantitative comparisons on GenEval.} To measure compositional generation capability, we conduct tests on GenEval, as shown in Tab. \ref{tab:geneval}. Our method surpasses both the base model and the BoN strategy across various dimensions. By leveraging the reward model for early evaluation and filtering, FR-TTS generates images where colors adhere to the prompt instructions more closely, while simultaneously showing small, yet consistent, improvements in all other aspects. To facilitate a comparison with ScalingAR, we also apply our strategy to LlamaGen. Our method similarly surpasses ScalingAR, which we attribute to the latter's over-reliance on endogenous signals.
\begin{table}[t]
\centering

\caption{\textbf{Ablations on reward calculation. We conduct ablation with Janus-pro 1B on TIIF-Bench.}}

\scalebox{0.87}{
\begin{tabular}{cccccc}
\toprule
Methods & Bas. & Adv. & Des. &Over. &Con.\\ 

Cropping &0.64 &0.50 &0.61 &0.54 &15s/img \\
ZeroPadding &0.63 &0.47 &0.62 &0.52 &15s/img\\
Complete Images &0.66 &0.51 &\textbf{0.67} &0.56 &24s/img \\
Filling-based Reward &\textbf{0.66} &\textbf{0.53} &0.66 &\textbf{0.57} &18s/img \\

\bottomrule
\end{tabular}}

\vspace{-8pt}
\label{tab:ablation_reward}

\end{table}
\paragraph{Quantitative comparisons on TIIF-Bench.} To evaluate the more comprehensive capabilities of both the model and the strategy, we conduct experiments on TIIF-Bench, as shown in Tab. \ref{tab:tiif}. Simple Importance Sampling (IS) struggles to outperform BoN on this benchmark, primarily because the reward model's evaluations are not accurate enough. Conversely, our method yields a substantial improvement across both the basic and advanced dimensions, with a more modest gain in the designer dimension. Viewing this from an alternative perspective, our method demonstrates a significantly greater improvement in long-text instruction following. This suggests that without the proactive intervention of an external reward model, the generation model finds it challenging to consistently adhere to instructions throughout the image synthesis process. Similarly, when comparing against ScalingAR on LlamaGen, our method achieves higher scores across three dimensions.
\vspace{-8pt}
\paragraph{visual comparisons.}We present a comparison of images generated w/ and w/o our FR-TTS in Fig. \ref{fig:visual_results}. The images produced by our strategy not only exhibit superior visual quality but also demonstrate enhanced instruction-following capability when dealing with longer texts, resulting in images with higher text fidelity.

\subsection{Ablation Study}

\paragraph{Ablations on reward calculation.} We conduct an ablation study on reward calculation methods in Tab. \ref{tab:ablation_reward}. We compare FR with Cropping, ZeroPadding, and complete images, as shown in Tab. \ref{tab:ablation_reward}. Even though Cropping and ZeroPadding have less consumption, they cannot surpass BoN. Meanwhile, using rewards of complete images just ties with our method, but has much more time consumption.
\begin{table}[h]
\centering
\begin{small}

\caption{\textbf{Ablations on method design. We conduct ablation with Janus-pro 1B on TIIF-Bench.}}

\begin{tabular}{cccc}
\toprule
Methods &Overall\\ 

w/o coarse-to-fine search strategy  &0.54  \\
w/o diversity reward  &0.55  \\
w/o dynamic weighting schedule  &0.51  \\

\bottomrule
\end{tabular}
\label{tab:ablation_design}
\end{small}
\end{table}
\begin{table}[h]
\centering

\caption{\textbf{Ablations on hyper-parameters. We conduct ablation with Janus-pro 1B on TIIF-Bench.}}
\scalebox{0.85}{
\begin{tabular}{cccccc}
\toprule
Block size & Filling times & Bas. & Adv. & Des. &Over.\\ 

1 &5 &0.62 &0.47 &0.59 &0.51 \\
6 &5 &0.64 &0.50 &0.61 &0.54\\
12&5 &0.66 &0.53 &0.66 &0.57 \\
24&5 &0.65 &0.51 &0.66 &0.55\\
12 &1 &0.64 &0.53 &0.62 &0.55\\
12 &10 &0.67 &0.56 &0.68 &0.59\\
12 &50 &0.68 &0.55 &0.66 &0.59\\

\bottomrule
\end{tabular}}

\label{tab:ablation_hyper}

\end{table}
\paragraph{Ablations on method design.} We conduct an ablation study on the different modules of FR-TTS in Tab. \ref{tab:ablation_design}, including the coarse-to-fine FR limit search strategy, diversity reward, and weighting schedule, as presented in Table 4. The coarse-to-fine search strategy yields a more accurate evaluation signal, while the diversity reward and weighting schedule mitigate the bias introduced by an FR that is inaccurate or lacks sufficient distinctiveness at certain steps.

\paragraph{Ablations on hyper-parameters.} We conduct an ablation experiment on two important hyperparameters within FR: block size and the number of random fillings, as presented in Tab. \ref{tab:ablation_hyper}. Primarily, an excessively large or small block size is detrimental to estimating the upper bound of the FR. An overly large block size leads to excessive image repetition and a lack of randomness, whereas an overly small block size introduces excessive randomness, thus increasing the difficulty of estimating the FR upper bound. 
Conversely, as the number of random fillings increases, the evaluation scores on the benchmark show continuous improvement, but this also results in higher consumption. We therefore need to achieve a trade-off between cost and performance.
\section{Conclusion}
\label{sec:conclusion}
In this work, we introduce FR-TTS, a novel test-time scaling strategy tailored for generation models under the NTP paradigm. The core of our approach lies in utilizing a filling-based reward as a robust evaluation signal for intermediate samples. Furthermore, we incorporate a Diversity reward to maintain the variety of early trajectories, employing a dynamic weighting schedule to effectively fuse these two signals. Our proposed strategy achieves superior performance across various reward models and established benchmarks. We hope this work encourages researchers to further explore the potential of test-time scaling.

{
    \small
    \bibliographystyle{ieeenat_fullname}
    \bibliography{main}

@article{chen2025tts,
  title={Tts-var: A test-time scaling framework for visual auto-regressive generation},
  author={Chen, Zhekai and Chu, Ruihang and Chen, Yukang and Zhang, Shiwei and Wei, Yujie and Zhang, Yingya and Liu, Xihui},
  journal={arXiv preprint arXiv:2507.18537},
  year={2025}
}

@article{singhcode,
  title={CoDe: Blockwise Control for Denoising Diffusion Models},
  author={Singh, Anuj and Mukherjee, Sayak and Beirami, Ahmad and Rad, Hadi J},
  journal={Transactions on Machine Learning Research}
}

@article{kim2025inference,
  title={Inference-time scaling for flow models via stochastic generation and rollover budget forcing},
  author={Kim, Jaihoon and Yoon, Taehoon and Hwang, Jisung and Sung, Minhyuk},
  journal={arXiv preprint arXiv:2503.19385},
  year={2025}
}

@inproceedings{ma2025scaling,
  title={Scaling Inference Time Compute for Diffusion Models},
  author={Ma, Nanye and Tong, Shangyuan and Jia, Haolin and Hu, Hexiang and Su, Yu-Chuan and Zhang, Mingda and Yang, Xuan and Li, Yandong and Jaakkola, Tommi and Jia, Xuhui and others},
  booktitle={Proceedings of the Computer Vision and Pattern Recognition Conference},
  pages={2523--2534},
  year={2025}
}

@article{tian2024visual,
  title={Visual autoregressive modeling: Scalable image generation via next-scale prediction},
  author={Tian, Keyu and Jiang, Yi and Yuan, Zehuan and Peng, Bingyue and Wang, Liwei},
  journal={Advances in neural information processing systems},
  volume={37},
  pages={84839--84865},
  year={2024}
}

@inproceedings{han2025infinity,
  title={Infinity: Scaling bitwise autoregressive modeling for high-resolution image synthesis},
  author={Han, Jian and Liu, Jinlai and Jiang, Yi and Yan, Bin and Zhang, Yuqi and Yuan, Zehuan and Peng, Bingyue and Liu, Xiaobing},
  booktitle={Proceedings of the Computer Vision and Pattern Recognition Conference},
  pages={15733--15744},
  year={2025}
}

@article{li2024autoregressive,
  title={Autoregressive image generation without vector quantization},
  author={Li, Tianhong and Tian, Yonglong and Li, He and Deng, Mingyang and He, Kaiming},
  journal={Advances in Neural Information Processing Systems},
  volume={37},
  pages={56424--56445},
  year={2024}
}

@article{chen2025go,
  title={Go with Your Gut: Scaling Confidence for Autoregressive Image Generation},
  author={Chen, Harold Haodong and Wu, Xianfeng and Shu, Wen-Jie and Guo, Rongjin and Lan, Disen and Yang, Harry and Chen, Ying-Cong},
  journal={arXiv preprint arXiv:2509.26376},
  year={2025}
}

@article{vaswani2017attention,
  title={Attention is all you need},
  author={Vaswani, Ashish and Shazeer, Noam and Parmar, Niki and Uszkoreit, Jakob and Jones, Llion and Gomez, Aidan N and Kaiser, {\L}ukasz and Polosukhin, Illia},
  journal={Advances in neural information processing systems},
  volume={30},
  year={2017}
}

@inproceedings{muennighoff2025s1,
  title={s1: Simple test-time scaling},
  author={Muennighoff, Niklas and Yang, Zitong and Shi, Weijia and Li, Xiang Lisa and Fei-Fei, Li and Hajishirzi, Hannaneh and Zettlemoyer, Luke and Liang, Percy and Cand{\`e}s, Emmanuel and Hashimoto, Tatsunori B},
  booktitle={Proceedings of the 2025 Conference on Empirical Methods in Natural Language Processing},
  pages={20286--20332},
  year={2025}
}

@article{snell2024scaling,
  title={Scaling llm test-time compute optimally can be more effective than scaling model parameters},
  author={Snell, Charlie and Lee, Jaehoon and Xu, Kelvin and Kumar, Aviral},
  journal={arXiv preprint arXiv:2408.03314},
  year={2024}
}

@article{wei2022chain,
  title={Chain-of-thought prompting elicits reasoning in large language models},
  author={Wei, Jason and Wang, Xuezhi and Schuurmans, Dale and Bosma, Maarten and Xia, Fei and Chi, Ed and Le, Quoc V and Zhou, Denny and others},
  journal={Advances in neural information processing systems},
  volume={35},
  pages={24824--24837},
  year={2022}
}

@article{feng2023towards,
  title={Towards revealing the mystery behind chain of thought: a theoretical perspective},
  author={Feng, Guhao and Zhang, Bohang and Gu, Yuntian and Ye, Haotian and He, Di and Wang, Liwei},
  journal={Advances in Neural Information Processing Systems},
  volume={36},
  pages={70757--70798},
  year={2023}
}

@article{li2024derivative,
  title={Derivative-free guidance in continuous and discrete diffusion models with soft value-based decoding},
  author={Li, Xiner and Zhao, Yulai and Wang, Chenyu and Scalia, Gabriele and Eraslan, Gokcen and Nair, Surag and Biancalani, Tommaso and Ji, Shuiwang and Regev, Aviv and Levine, Sergey and others},
  journal={arXiv preprint arXiv:2408.08252},
  year={2024}
}

@inproceedings{esser2024scaling,
  title={Scaling rectified flow transformers for high-resolution image synthesis},
  author={Esser, Patrick and Kulal, Sumith and Blattmann, Andreas and Entezari, Rahim and M{\"u}ller, Jonas and Saini, Harry and Levi, Yam and Lorenz, Dominik and Sauer, Axel and Boesel, Frederic and others},
  booktitle={Forty-first international conference on machine learning},
  year={2024}
}

@inproceedings{lee2022autoregressive,
  title={Autoregressive image generation using residual quantization},
  author={Lee, Doyup and Kim, Chiheon and Kim, Saehoon and Cho, Minsu and Han, Wook-Shin},
  booktitle={Proceedings of the IEEE/CVF conference on computer vision and pattern recognition},
  pages={11523--11532},
  year={2022}
}

@article{xu2023imagereward,
  title={Imagereward: Learning and evaluating human preferences for text-to-image generation},
  author={Xu, Jiazheng and Liu, Xiao and Wu, Yuchen and Tong, Yuxuan and Li, Qinkai and Ding, Ming and Tang, Jie and Dong, Yuxiao},
  journal={Advances in Neural Information Processing Systems},
  volume={36},
  pages={15903--15935},
  year={2023}
}

@inproceedings{ma2025hpsv3,
  title={Hpsv3: Towards wide-spectrum human preference score},
  author={Ma, Yuhang and Wu, Xiaoshi and Sun, Keqiang and Li, Hongsheng},
  booktitle={Proceedings of the IEEE/CVF International Conference on Computer Vision},
  pages={15086--15095},
  year={2025}
}

@article{wu2023human,
  title={Human preference score v2: A solid benchmark for evaluating human preferences of text-to-image synthesis},
  author={Wu, Xiaoshi and Hao, Yiming and Sun, Keqiang and Chen, Yixiong and Zhu, Feng and Zhao, Rui and Li, Hongsheng},
  journal={arXiv preprint arXiv:2306.09341},
  year={2023}
}

@article{wissler1905spearman,
  title={The Spearman correlation formula},
  author={Wissler, Clark},
  journal={Science},
  volume={22},
  number={558},
  pages={309--311},
  year={1905},
  publisher={American Association for the Advancement of Science}
}

@article{chen2025janus,
  title={Janus-pro: Unified multimodal understanding and generation with data and model scaling},
  author={Chen, Xiaokang and Wu, Zhiyu and Liu, Xingchao and Pan, Zizheng and Liu, Wen and Xie, Zhenda and Yu, Xingkai and Ruan, Chong},
  journal={arXiv preprint arXiv:2501.17811},
  year={2025}
}

@misc{OpenImagePreferencesV1Blog, author = {{Data Is Better Together}}, title = {{Open Preference Dataset for Text-to-Image Generation by the Community}}, howpublished = {Hugging Face Blog}, year = {2024}, note = {Available at: \url{https://huggingface.co/blog/image-preferences}} }

@article{ma2025towards,
  title={Towards Better \& Faster Autoregressive Image Generation: From the Perspective of Entropy},
  author={Ma, Xiaoxiao and Zhao, Feng and Ling, Pengyang and Qiu, Haibo and Wei, Zhixiang and Yu, Hu and Huang, Jie and Zeng, Zhixiong and Ma, Lin},
  journal={arXiv preprint arXiv:2510.09012},
  year={2025}
}

@misc{flux2024,
    author={Black Forest Labs},
    title={FLUX},
    year={2024},
    howpublished={\url{https://github.com/black-forest-labs/flux}},
}

@article{yue2025understand,
  title={Understand Before You Generate: Self-Guided Training for Autoregressive Image Generation},
  author={Yue, Xiaoyu and Wang, Zidong and Wang, Yuqing and Zhang, Wenlong and Liu, Xihui and Ouyang, Wanli and Bai, Lei and Zhou, Luping},
  journal={arXiv preprint arXiv:2509.15185},
  year={2025}
}

@incollection{efros2023image,
  title={Image quilting for texture synthesis and transfer},
  author={Efros, Alexei A and Freeman, William T},
  booktitle={Seminal graphics papers: pushing the boundaries, volume 2},
  pages={571--576},
  year={2023}
}

@article{barnes2009patchmatch,
  title={PatchMatch: A randomized correspondence algorithm for structural image editing},
  author={Barnes, Connelly and Shechtman, Eli and Finkelstein, Adam and Goldman, Dan B},
  journal={ACM Trans. Graph.},
  volume={28},
  number={3},
  pages={24},
  year={2009}
}

@article{sun2024autoregressive,
  title={Autoregressive model beats diffusion: Llama for scalable image generation},
  author={Sun, Peize and Jiang, Yi and Chen, Shoufa and Zhang, Shilong and Peng, Bingyue and Luo, Ping and Yuan, Zehuan},
  journal={arXiv preprint arXiv:2406.06525},
  year={2024}
}

@article{podell2023sdxl,
  title={Sdxl: Improving latent diffusion models for high-resolution image synthesis},
  author={Podell, Dustin and English, Zion and Lacey, Kyle and Blattmann, Andreas and Dockhorn, Tim and M{\"u}ller, Jonas and Penna, Joe and Rombach, Robin},
  journal={arXiv preprint arXiv:2307.01952},
  year={2023}
}

@article{wu2025lightgen,
  title={Lightgen: Efficient image generation through knowledge distillation and direct preference optimization},
  author={Wu, Xianfeng and Bai, Yajing and Zheng, Haoze and Chen, Harold Haodong and Liu, Yexin and Wang, Zihao and Ma, Xuran and Shu, Wen-Jie and Wu, Xianzu and Yang, Harry and others},
  journal={arXiv preprint arXiv:2503.08619},
  year={2025}
}

@article{wang2024emu3,
  title={Emu3: Next-token prediction is all you need},
  author={Wang, Xinlong and Zhang, Xiaosong and Luo, Zhengxiong and Sun, Quan and Cui, Yufeng and Wang, Jinsheng and Zhang, Fan and Wang, Yueze and Li, Zhen and Yu, Qiying and others},
  journal={arXiv preprint arXiv:2409.18869},
  year={2024}
}

@article{xie2024show,
  title={Show-o: One single transformer to unify multimodal understanding and generation},
  author={Xie, Jinheng and Mao, Weijia and Bai, Zechen and Zhang, David Junhao and Wang, Weihao and Lin, Kevin Qinghong and Gu, Yuchao and Chen, Zhijie and Yang, Zhenheng and Shou, Mike Zheng},
  journal={arXiv preprint arXiv:2408.12528},
  year={2024}
}

@inproceedings{hessel2021clipscore,
  title={Clipscore: A reference-free evaluation metric for image captioning},
  author={Hessel, Jack and Holtzman, Ari and Forbes, Maxwell and Le Bras, Ronan and Choi, Yejin},
  booktitle={Proceedings of the 2021 conference on empirical methods in natural language processing},
  pages={7514--7528},
  year={2021}
}

@misc{Schuhmann:aesthetics,
    title        = {{Laion aesthetics}},
    author       = {C. Schuhmann},
    year={2022},
    howpublished={\url{https://laion.ai/blog/ laion- aesthetics}},
}

@article{ghosh2023geneval,
  title={Geneval: An object-focused framework for evaluating text-to-image alignment},
  author={Ghosh, Dhruba and Hajishirzi, Hannaneh and Schmidt, Ludwig},
  journal={Advances in Neural Information Processing Systems},
  volume={36},
  pages={52132--52152},
  year={2023}
}

@article{wei2025tiif,
  title={TIIF-Bench: How Does Your T2I Model Follow Your Instructions?},
  author={Wei, Xinyu and Zhang, Jinrui and Wang, Zeqing and Wei, Hongyang and Guo, Zhen and Zhang, Lei},
  journal={arXiv preprint arXiv:2506.02161},
  year={2025}
}

@article{simonyan2014very,
  title={Very deep convolutional networks for large-scale image recognition},
  author={Simonyan, Karen and Zisserman, Andrew},
  journal={arXiv preprint arXiv:1409.1556},
  year={2014}
}
}


\end{document}